\documentclass[acmsmall,nonacm]{acmart}
\usepackage{mathtools}
\usepackage{booktabs,multirow,array,tabularx,threeparttable}
\usepackage{graphicx,adjustbox,placeins}
\usepackage{tikz}
\usetikzlibrary{arrows.meta,positioning,calc}
\usepackage{microtype,xurl}
\usepackage{xcolor}
\definecolor{tcfaaccent}{RGB}{30,82,132}
\definecolor{tcfagreen}{RGB}{29,115,89}
\definecolor{tcfagrey}{RGB}{110,110,110}

\newcommand{\Yes}{\ensuremath{\mathrm{Yes}}}
\newcommand{\No}{\ensuremath{\mathrm{No}}}
\newcommand{\GR}{\ensuremath{\mathrm{GR}}}
\newcommand{\IR}{\ensuremath{\mathrm{IR}}}

\newcommand{\Ind}{\mathbb{I}}

\newcommand{\Ncand}{\mathcal{N}}

\newcommand{\Loss}{\mathcal{L}}

\newcommand{\pp}{\,\mathrm{pp}}

\makeatletter
\let\datainput\@@input
\makeatother
\newcommand{\tabnote}[1]{\par\vspace{2pt}{\footnotesize\emph{Note.} #1}}
\acmJournal{TOMM}
\setcopyright{none}
\acmDOI{}
\begin{document}
\title{Third-Count False Acceptance in Small-Parameter VLMs: Diagnosis, Calibration, and Recovery Limits}
\renewcommand{\shorttitle}{Third-Count False Acceptance in Small-Parameter VLMs}
\author{Shengqi Zhou}
\affiliation{\institution{Northeast Petroleum University}\city{Daqing}\country{China}}
\email{shbiao@stu.nepu.edu.cn}
\author{Wenxiang Zhu}
\affiliation{\institution{Tongji University}\city{Shanghai}\country{China}}
\author{Zhentao Cang}
\affiliation{\institution{Northwestern Polytechnical University}\city{Xi'an}\country{China}}
\author{Yuhan Zhu}
\affiliation{\institution{Northeast Petroleum University}\city{Daqing}\country{China}}
\begin{abstract}
A small-parameter vision--language model can miscount an image and independently endorse a third quantity that differs from both its open count and the annotation. We diagnose this error, \emph{third-count false acceptance} (TCFA), through five relations among annotated, open, and proposed counts. Across six open-weight checkpoints in four model families, nearly identical balanced accuracy can conceal sharply different acceptance profiles. We measure paired recovery (\(J_{43}\)) as joint acceptance of the annotation and rejection of the frozen Base counting error on the same image. On COCO-ReM Dev, calibrating Gemma-3 \(q/v\) (seed~0) standard continuation at a 20\% false-acceptance budget raises paired recovery from 26 to 58 of 318 images without changing scores. Across 24 paired evaluations spanning COCO-ReM Test, external TallyQA Test, two 4B models, three seeds, and two readouts, hard-positive continuation raises true acceptance, false acceptance, and third-count acceptance in every run. Under default generated decisions, paired recovery changes by \(+3.62\) percentage points for Qwen3.5 and \(-5.69\) for Gemma-3 on COCO-ReM Test (with false acceptance up by 4.06 and 26.16 points), and by \(-0.53\) and \(-14.50\) points on TallyQA Test (with false acceptance up by 1.68 and 15.47 points). Transferring locked Dev 20\% FAR thresholds to seed-0 Test sequence scores eliminates Qwen3.5's COCO-ReM advantage (\(-0.19\) points at \(\approx 18\%\) FAR) while raising Gemma-3's TallyQA recovery on standard continuation by \(+21.08\) points (\(25\to 103\) of 370). Even among images passing \(J_{43}\), 36\%--62\% after calibration (and 43\%--87\% under default generation) still endorse at least one third wrong count, showing why paired recovery and TCFA must be evaluated alongside false acceptance.
\end{abstract}

\ccsdesc[500]{Computing methodologies~Computer vision}
\keywords{vision--language models, visual counting, numerical premise verification, third-count false acceptance, relation-conditioned training}
\maketitle

\section{Introduction}\label{sec:intro}
A multimedia system often receives a quantity from one component and checks it against an image in another. Caption auditing evaluates a count written in generated or editorial text. Inventory review verifies an item record against a photograph. Image retrieval applies numeric predicates such as \emph{exactly two people} to otherwise relevant images. In these workflows, a false approval retains an incorrect description, record, or retrieval result. A numerical pipeline therefore needs accurate counts and reliable rejection of false premises. We focus on small-parameter, open-weight VLMs in the nominal 3B--4B class, a practical setting for local image verification~\cite{smolvlm}.

\begin{figure}[t]
\centering
\includegraphics[width=\textwidth]{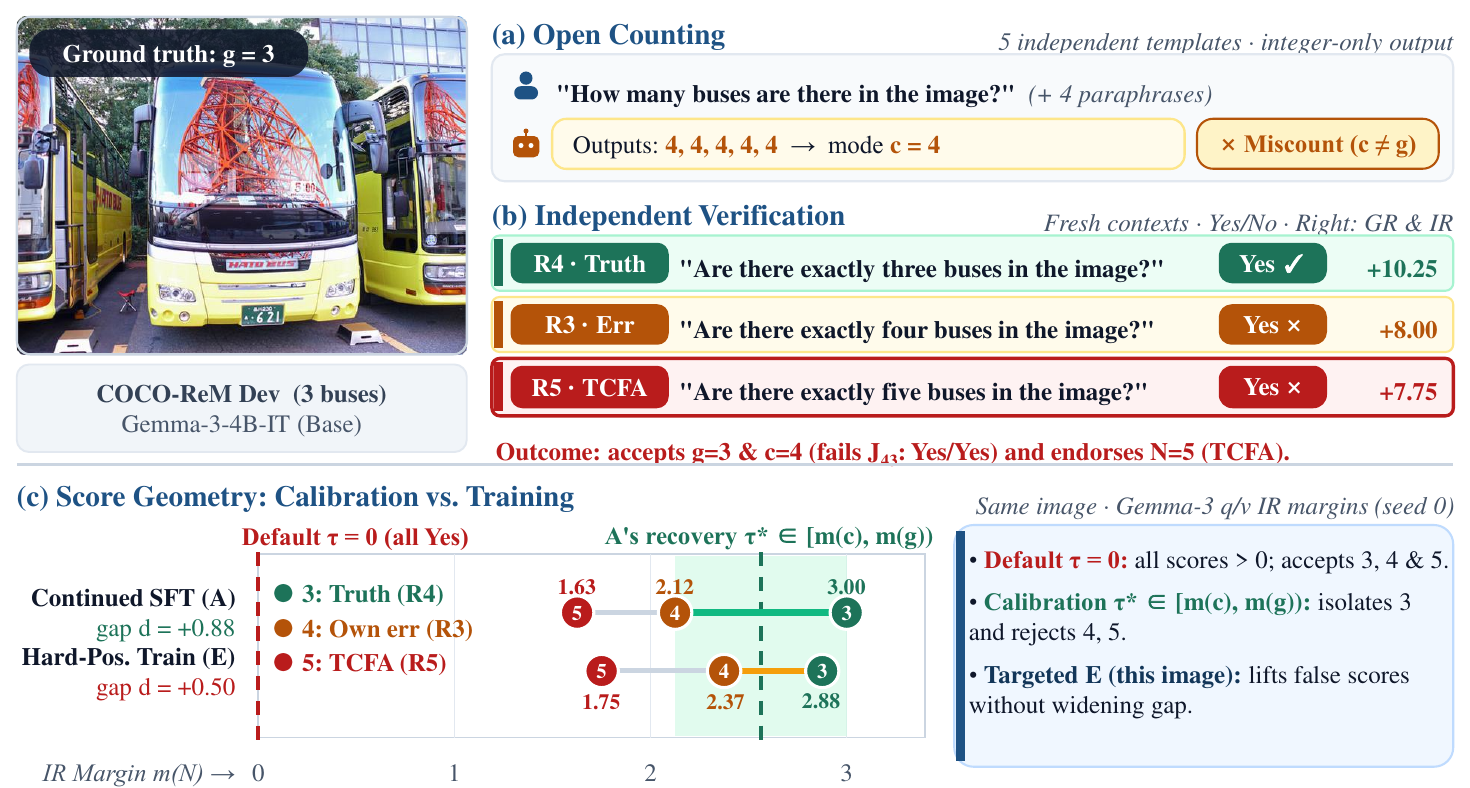}
\caption{Overview of third-count false acceptance (TCFA), paired recovery, and score geometry on a COCO-ReM Dev image (\(g=3\) buses; Gemma-3-4B-IT). \textbf{(a)}~Five independent open-count templates yield a unique miscount mode \(c=4\). \textbf{(b)}~In fresh verification contexts, the Base model answers \Yes{} to \(N=3\) (R4), \(N=4\) (R3), and \(N=5\) (R5, TCFA), failing paired recovery \(J_{43}\) via a \Yes/\Yes{} pair. \textbf{(c)}~On the same image under aligned \(q/v\) adapters (seed~0), default \(\tau=0\) accepts all three premises, whereas calibrating the threshold into the shaded recovery band \(\tau^*\in[m(c),m(g))\) for standard continuation (A) isolates the annotation and rejects both false counts; targeted continuation (E) raises both false-premise margins without widening the pair gap on this image.}
\Description{A three-panel overview figure showing a photograph of three yellow buses from COCO-ReM Dev alongside Gemma-3-4B-IT outputs. Panel a shows five independent open counting prompts all returning 4, giving mode c equals 4. Panel b shows independent Yes/No verification answering Yes to three, four, and five buses with positive IR margins, illustrating R4 truth, R3 own error, and R5 third-count false acceptance. Panel c compares IR margins for standard continuation A and targeted continuation E on the same image, showing a shaded recovery threshold band between the error and truth scores for A where calibration isolates the true count three.}
\label{fig:three}
\end{figure}

The counting and verification interfaces can make different decisions about the same image. Given a photograph containing three buses (Figure~\ref{fig:three}), a VLM can answer four to an open counting question and still answer Yes when independently asked whether there are five. The number five is neither the annotated count nor the VLM's own open answer, yet a downstream system would accept it as visually supported. Existing hallucination analyses establish number errors, sensitivity to misleading premises, and inconsistencies across related tasks~\cite{quantity,pih,pvfh}. In independent contexts, we study the relation among the annotated count \(g\), the unique mode \(c\) from five open-count prompts, and a proposed count \(N\). Their five relations separate correct agreement, false-premise acceptance after a correct count, endorsement of the model's own error, recovery of the annotation after a miscount, and acceptance of a third number.

On COCO-ReM Dev, Base TCFA varies sharply across families---rare in conservative Qwen3.5 (1.76\% under GR) yet pervasive in Gemma-3 (43.16\%) and Phi-3.5 (51.93\%)---and Qwen3.5 and Phi-3.5 attain nearly identical balanced accuracy despite opposing acceptance profiles. Joint SFT improves Qwen3.5 open counting while raising both false approvals and TCFA on the same questions. In both cases, aggregate accuracy hides how the verifier handles competing quantities on the same image. We therefore pair acceptance of the annotation with rejection of the frozen Base counting error to test whether a verification system makes a usable correction.

A verifier can recover more images when its threshold moves into the interval between the two scores, even with the same ordering. The center of the interval describes how readily the verifier accepts both premises, and its width describes how far apart their scores are. On Dev, threshold calibration alone lifts Gemma-3 \(q/v\) recovery dramatically without altering scores. For independent testing, we match \(q/v\) adapter scope across three paired seeds for Qwen3.5-4B and Gemma-3-4B-IT on COCO-ReM Test and external TallyQA Test. Hard-positive continuation raises true acceptance, false acceptance, and third-count acceptance in all 24 paired runs, yet under default decisions produces positive point estimates in all three seeds only for Qwen3.5 on COCO-ReM, while reducing Gemma-3 recovery on both benchmarks and eliminating Qwen3.5's gain on TallyQA. Applying locked Dev 20\% FAR thresholds to seed-0 Test sequence scores erases Qwen3.5's COCO-ReM advantage at matched risk and produces far larger recovery gains on Gemma-3 than targeted retraining.

\paragraph{Contributions}
\begin{enumerate}
\item \textbf{Five-relation diagnosis.} We partition numerical verification by the relation among the annotated count, open count, and proposed count. Across six checkpoints from four VLM families, the partition separates wrong-count agreement (R3), truth recovery (R4), and TCFA (R5), and shows that even recovered images frequently endorse a third wrong count.
\item \textbf{Paired truth recovery and risk-controlled calibration.} Using a fixed Base-error reference, we pair annotation acceptance with rejection of the original error (\(J_{43}\)) and test score-threshold calibration both out-of-fold on Dev and under locked Dev-to-Test transfer.
\item \textbf{Dual-benchmark training comparison.} Across COCO-ReM Test and simple Visual Genome category-count questions from TallyQA Test, targeted continuation uniformly raises TAR, FAR, and TCFA across all 24 runs while leaving same-image score ordering nearly unchanged and yielding no net gain under matched Dev-locked risk.
\end{enumerate}

\section{Related work}\label{sec:related}
\subsection{Multimodal hallucination and false premises}
Multimedia verification must distinguish a plausible description from one supported by an image. CHAIR measures unsupported objects in captions, and POPE asks whether an object is present~\cite{chair,pope}. Hallu-PI compares responses under image and prompt perturbations, PhD varies the context supplied with an image, and HallusionBench distinguishes errors associated with language hallucinations and visual illusions~\cite{ding2024hallupi,phd,hallusionbench}. DASH evaluates systematic false claims about object presence across many images~\cite{dash}. In numerical verification, competing premises can all name an object that is present. The question is which exact quantity the image supports.

Complementary-image VQA and Possible Worlds VQA show how visual evidence and language priors can change answers to related questions~\cite{goyal2017vqa,vosoughi2024pwvqa}. \emph{Words or Vision} tests conflicts between text and image evidence, and FINER varies fine-grained object, attribute, and relation claims~\cite{words,xiao2026finer}. Other benchmarks vary the supplied numerical premise, visual distractors, or fine-grained image differences~\cite{pih,pvfh,weng2026oddgrid}. NegBench tests whether wording that excludes a concept changes retrieval and discrimination~\cite{negbench}. We vary the proposed number with the image and category held fixed and record whether the model endorses its own counting error or another wrong quantity.

Existing interventions act at distinct points in the decision pipeline. VCD contrasts decoding scores under original and distorted images, OPERA changes token selection through an over-trust penalty, and LURE revises a generated description~\cite{vcd,opera,lure}. DPO learns from response preferences, and VGL-DPO weights the preference objective using visual grounding at the token level~\cite{rafailov2023dpo,li2026vgldpo}. Analysis of attention and correctness also distinguishes access to visual evidence from a correct answer~\cite{vea}. We calibrate the decision threshold on fixed answer-sequence scores and separately test how supervision changes the scores.

\subsection{Visual counting and cross-task consistency}
Visual counting research evaluates both category counts and counts defined by language. TallyQA benchmarks counting questions with attributes and relations. CLIP-Count estimates density for a text-described target, and CountBench evaluates count-sensitive image--text matching~\cite{acharya2019tallyqa,jiang2023clipcount,paiss2023countbench}. CounterCount varies factual and counterfactual attribute counts to probe numerical priors. \emph{Counting Circuits} uses activation patching and head-level analysis to study counting mechanisms~\cite{countercount,circuits}. We pair open counting with independent verification to measure which numerical premises a verifier approves after an observed counting error.

\emph{Quantity Matters} is the closest precedent~\cite{quantity}. It tests verification of the ground truth, the model's generated count, and nearby quantities. It documents acceptance of mutually contradictory premises and trains more consistent counting, verification, and comparison responses. We freeze the Base counting errors across adapted models, partition conditions by five numerical relations, and require a correct joint decision on the annotated count and the specific Base error. The fixed reference keeps the difficulty cohort stable after adaptation. Reporting paired recovery alongside false acceptance measures the benefit and cost of correction. The resulting evaluation distinguishes agreement that preserves an error from disagreement that corrects it.

\subsection{Score calibration and risk-constrained decisions}
Probability calibration aligns confidence estimates with empirical accuracy~\cite{guo2017calibration}. Selective prediction uses confidence to trade off prediction coverage and error among retained cases~\cite{geifman2017selective}. Neyman--Pearson classification studies decision rules under a bound on false positives~\cite{tong2018np}. Our calibration is a threshold selection procedure using validation scores. Every valid score receives a binary decision. The false-acceptance budget applies to the common false-premise pool, and threshold selection maximizes correct joint decisions on a fixed Base-error cohort. On Dev we report both the requested budget and the realized FAR across the held-out folds.

\subsection{Hard examples and group-weighted training}
Hard-example weighting focuses supervision on examples a model finds difficult~\cite{lin2017focal}. Our hard-positive objective selects true premises explicitly rejected by the frozen Base under generated verification. The relation-conditioned objective selects true premises from images with erroneous Base open counts. Group DRO raises the weight of groups with high training loss~\cite{sagawa2020groupdro}. Applying all three choices to the same verification relations separates the effect of selecting recovery examples from more general difficulty-based reweighting.

\section{Three-quantity analysis framework}\label{sec:framework}
\subsection{Images, targets, and independent decisions}
For an evaluated unit $i$, let $I_i$ be an image, $s_i$ an exact-count target expression, and $g_i\in\mathbb{Z}_{\geq 0}$ its annotated count. On COCO-ReM, $s_i$ denotes an annotated object category. The TallyQA protocol preserves the target phrase from the original question.

Each unit is queried with five independently instantiated open-count prompts. Their parsed integer answers are $a_i^{(1)},\ldots,a_i^{(5)}$. If all are valid and the empirical mode is unique, let
\begin{equation}\label{eq:mode}
 c_i=\underset{n\in\mathbb{Z}_{\ge0}}{\arg\max}\,\sum_{j=1}^{5}\Ind[a_i^{(j)}=n].
\end{equation}
A tied mode or an invalid response leaves $c_i$ undefined. We report the fraction of images with a defined mode.

Using several phrasings reduces dependence on one counting instruction. A unique mode captures the most frequent answer when alternatives split across several values. Coverage records how often the protocol yields this reference. Fresh verification contexts mirror a system that receives the numerical claim separately and keep the counting response out of the verifier's prompt.

A fresh verification context presents the image and asks whether it contains exactly $N$ instances of $s_i$. The readout-specific response is $y_{i,N}^{(r)}\in\{\Yes,\No,\bot\}$, where $r\in\{\GR,\IR\}$ and $\bot$ means invalid or unresolved. The correct response is \Yes{} exactly when $N=g_i$.

\subsection{Common verification candidates}\label{sec:candidates}
To prevent a model-specific open count from changing the primary task, the common quantity plan is centered on the ground truth:
\begin{equation}\label{eq:window}
 \Ncand_i=\{n\in\mathbb{Z}_{\ge0}:g_i-2\le n\le g_i+2\}.
\end{equation}
This produces five quantities for $g_i\ge2$ and four for $g_i=1$. The 500 COCO-ReM development units give 2,450 shared primary verification conditions: 500 true premises and 1,950 false premises. The formal Test candidate set is the primary window plus \(N=c_i^{(0)}\) when the Base mode is defined. Each adapted model also answers the five open-count prompts to measure its counting accuracy. Overall TAR, FAR, BA, and fixed-reference TCFA use the primary window. The supplementary Base-count question extends coverage of the paired endpoint when the Base error falls outside that window.

The centered window tests nearby errors of equal magnitude wherever the count range allows both signs. One- and two-instance discrepancies are plausible alternatives for caption and inventory checks. The shared window gives arms the same questions and false-premise prevalence. Supplementing it with the actual open-count error retains both sides of the recovery test on images miscounted by more than two.

\subsection{Five mutually exclusive numerical relationships}
For $c_i$ defined, Table~\ref{tab:relations} partitions every numerical premise into five disjoint groups. In particular, R5 is the candidate population for TCFA, whereas R4 is a positive premise despite erroneous open counting.
\begin{table}[!tbp]
\caption{Joint three-quantity relationships. ``Yes'' is the desired verification answer only for R1 and R4. An observed \Yes{} in R5 constitutes TCFA.}
\label{tab:relations}
\centering\footnotesize\setlength{\tabcolsep}{3pt}
\begin{tabular}{@{}cllc@{}}
\toprule
Group & Condition & Meaning of \Yes & Target \\
\midrule
R1 & $c=g,\ N=g$ & correct agreement & \Yes \\
R2 & $c=g,\ N\ne g$ & false acceptance & \No \\
R3 & $c\ne g,\ N=c$ & wrong-count agreement & \No \\
R4 & $c\ne g,\ N=g$ & truth recovery & \Yes \\
R5 & $c\ne g,\ N\notin\{c,g\}$ & third-count acceptance & \No \\
\bottomrule
\end{tabular}
\end{table}

Define the eligible third-count indicator
\begin{equation}
 q_{i,N}=\Ind[c_i\text{ defined},\ c_i\ne g_i,\ N\ne c_i,\ N\ne g_i].
\end{equation}
The TCFA event is $q_{i,N}\Ind[y_{i,N}^{(r)}=\Yes]$. Figure~\ref{fig:three}(a)--(b) illustrates the three relations (R3--R5) for an image with $g=3$ and $c=4$.

\subsection{Frozen-reference and model-specific analyses}
Frozen-reference analysis uses the same Base mode \(c_i^{(0)}\) for every trained arm, which fixes the relation groups for paired comparisons. Model-specific analysis uses each adapted model's own mode \(c_i^{(h)}\) to describe its current decision profile and TCFA rate.

\section{Targeted recovery training}\label{sec:method}
We compare standard continued SFT (A) with an objective that increases supervision on true claims rejected by the frozen Base verifier (E). Both arms use the same image--target units and start from the same model-specific Joint SFT checkpoint. Frozen Base counts assign relation membership. Arm D adds supervision tied to Base counting errors. Appendix~\ref{app:legacy} defines the controls for relation balancing, positive-loss protection, and group-weighted training.

\subsection{Training units and sequence supervision}
Each grouped update combines a Base-correct label, which supplies R1 and R2, and a Base-incorrect label, which supplies R3--R5. Each label also supplies one counting question. R2 and R5 false premises come from the ground-truth window, with R5 distinct from the Base count. R3 uses \(N=c_i^{(0)}\) even outside the window. On 2,500 training images, Qwen3.5 has 798 correct Base modes, 1,659 wrong modes, and 43 undefined modes. We cycle the smaller defined stratum and use counting-only updates for undefined modes.

For question \(x_i\), we supervise the complete target answer \(w_i\), including its native end-of-sequence token, with mean token cross-entropy
\begin{equation}\label{eq:seqce}
 \ell_{\mathrm{seq}}(x_i,w_i;\theta)=-\frac{1}{|w_i|}\sum_{t=1}^{|w_i|}\log p_\theta(w_{i,t}\mid x_i,w_{i,<t}).
\end{equation}
Let \(L_k\) be the sequence CE for relation R\(k\) and \(L_{\mathrm{count}}\) the mean loss on the counting prompts. In standard continuation (A), true and false premises receive equal total weight.
\begin{equation}
 \Loss_A=L_{\mathrm{count}}+\tfrac14(L_1+L_4)+\tfrac16(L_2+L_3+L_5).
\end{equation}
The targeted objectives weight the five relations equally and protect the two positive relations. With \(L_k^{\mathrm{ref}}\) denoting positive-prompt CE under the Joint SFT initialization, define
\begin{equation}
 P=[L_1-L_1^{\mathrm{ref}}]_+ + [L_4-L_4^{\mathrm{ref}}]_+ ,
 \qquad [z]_+=\max(z,0).
\end{equation}
Arm D adds supervision to R4, whose true premise is paired with a Base counting error. Arm E selects hard positives that frozen Base answers No under GR on Train. Let \(H_k\) indicate a selected positive for relation \(k\). The objectives are
\begin{align}
 \Loss_D &= L_{\mathrm{count}}+\tfrac15\sum_{k=1}^{5}L_k +P+0.2L_4,\\
 \Loss_E &= L_{\mathrm{count}}+\tfrac15\sum_{k=1}^{5}L_k +P \notag\\
         &\quad + \eta_E(H_1L_1+H_4L_4).
\end{align}
We set \(\eta_E=0.2n_4/n_E\) using the numbers of R4 and selected hard-positive occurrences in grouped updates, matching the total extra coefficient in D and E. Selecting hard positives for E requires one additional frozen-Base GR verification pass over the 2,500 Train true premises, whereas D relies only on Base open counts. Arm D serves as a Dev ablation, while the formal dual-Test comparison evaluates the full targeted objective \(\Loss_E\) against standard continuation \(\Loss_A\) (with relation weighting and positive protection isolated separately on Dev in Appendix~\ref{app:legacy}). Counting-only updates use \(L_{\mathrm{count}}\) throughout.

\section{Evaluation design}\label{sec:design}
\subsection{Datasets and models}
\begin{table}[!htbp]
\caption{COCO-ReM partitions and the external TallyQA selection. Each image contributes one target count.}
\label{tab:data}
\centering\small\setlength{\tabcolsep}{3pt}
\begin{tabular}{@{}llrl@{}}
\toprule
Source & Split & Images & Use \\
\midrule
COCO-ReM & Train & 2,500 & Training \\
COCO-ReM & Dev & 500 & Development \\
COCO-ReM & Test & 800 & Holdout \\
TallyQA & Original test & 800 & External evaluation \\
\bottomrule
\end{tabular}
\end{table}

We construct exact-count labels from the refined instance annotations in COCO-ReM~\cite{cocorem}. Its annotation review corrects incomplete instances, mislabeled masks, and imprecise boundaries found in COCO 2017~\cite{lin2014coco}. The image-disjoint splits each use one image--category pair per image and contain 10\% of labels at count 1, 10\% at count 2, and 80\% at counts 3--12. The split sizes appear in Table~\ref{tab:data}. We use a fixed random seed for sampling.

The external TallyQA evaluation samples 800 images and 800 corresponding questions from the original TallyQA test split~\cite{acharya2019tallyqa}, with 120 labels each at counts 1 and 2 and 560 at counts 3--12. Starting from 38,589 raw test questions, we exclude 15,598 complex questions, 15,505 non-category or out-of-vocabulary questions, 176 outside counts 1--12, and 9 conflicting image--subject pairs, leaving 7,131 simple category-count labels over 6,886 images. All 800 sampled images come from Visual Genome (447 from \texttt{VG\_100K} and 353 from \texttt{VG\_100K\_2}), contain exclusively simple category-count questions, and have zero overlap with COCO-ReM by COCO image ID and exact file deduplication.

The Base diagnosis covers six open-weight checkpoints in the nominal 3B--4B class on COCO-ReM Dev. We evaluate Qwen2.5-VL-3B~\cite{qwen25vl}, Qwen3-VL-4B~\cite{qwen3vl}, Qwen3.5-4B~\cite{qwen35}, Phi-3.5-Vision-Instruct~\cite{phi35vision}, Gemma-3-4B-IT~\cite{gemma3}, and LFM2.5-VL-3B~\cite{lfm25vl}. The nominal sizes follow the model names or release descriptions. Exact totals vary with the vision encoder. The independent A/E comparison uses Qwen3.5-4B and Gemma-3-4B-IT with the same \(q/v\) adapter scope. Gemma's \(q/k/v\) Dev experiment assesses sensitivity to the wider adapter configuration.

\subsection{Open counting and verification readouts}\label{sec:readouts}
Each open-count prompt requests one nonnegative integer in digits. We trim whitespace, require a full ASCII-decimal match, and require native termination. The five responses define the reference for both verification readouts. The counting prompts are listed below.

\textbf{Generated readout (GR)} requests a single Yes or No. A completed response is valid when its trimmed text matches either label after case normalization. \textbf{Sequence-scored readout (IR)} appends the assistant prefix \texttt{FINAL\_ANSWER:} and sums token log-probabilities for the contextualized Yes and No sequences, each with its designated native terminating token. Their log-probability difference \(m_{i,N}\) is the verification margin. Positive and negative margins select Yes and No, respectively. The default IR parser records a zero margin as an unresolved tie.

GR evaluates the answer an application receives from the model. IR compares two complete answers under a common prefix, including their termination probability. Summing contextual token log-probabilities accounts for answers that span more than one token and provides a margin usable for calibration. Qwen3.5 uses BF16, eager attention, greedy generation at seed 0, disabled thinking, a 2,048-token maximum, and a 120-second per-question limit.

\paragraph{Counting prompts}\label{sec:counting-prompts}
The five prompts use the plural target phrase in the slot \texttt{\{subjects\}}.
\begin{enumerate}
\item How many \texttt{\{subjects\}} are there in the image?
\item How many \texttt{\{subjects\}} can you see in this picture?
\item What is the total number of \texttt{\{subjects\}} shown in the image?
\item Count the \texttt{\{subjects\}} visible in this picture. How many are there?
\item Looking at this image, how many \texttt{\{subjects\}} are visible?
\end{enumerate}
Each template requests a single nonnegative integer in digits. Verification asks whether the image contains exactly the proposed quantity and requests a Yes or No answer. Designated answer EOS tokens are 248044 (Qwen3.5), 106 (Gemma-3), 32000 (Phi-3.5), and 124900 (LFM2.5). On TallyQA, the original question replaces the first template and the remaining four preserve its target phrase.

\subsection{Performance, coverage, and uncertainty}\label{sec:metrics}
Let $V_+$ and $V_-$ denote valid true-premise and false-premise questions in the primary window, and let $V=V_+\cup V_-$. We use
\begin{align}
 \mathrm{TAR}&=\frac{\sum_{(i,N)\in V_+}\Ind[y_{i,N}=\Yes]}{|V_+|},\label{eq:tar}\\
 \mathrm{FAR}&=\frac{\sum_{(i,N)\in V_-}\Ind[y_{i,N}=\Yes]}{|V_-|},\label{eq:far}\\
 \mathrm{BA}&=\tfrac12\bigl(\mathrm{TAR}+1-\mathrm{FAR}\bigr),\label{eq:ba}\\
 \mathrm{TCFA}&=\frac{\sum_{(i,N)\in V}q_{i,N}\Ind[y_{i,N}=\Yes]}{\sum_{(i,N)\in V}q_{i,N}}.\label{eq:tcfa}
\end{align}
Counting accuracy uses all images, mode coverage counts defined modes, and relation-specific rates use valid answers in each group. Let \(\mathcal W_{a,b}\) contain images with defined \(c_i^{(0)}\ne g_i\) and valid responses to both premises in both compared arms. With \(u_i^{(r,h)}\) denoting acceptance of \(g_i\) and \(v_i^{(r,h)}\) rejection of \(c_i^{(0)}\), we use
\begin{equation}\label{eq:j43}
 J_{43}^{(r)}(h)=\frac{1}{|\mathcal W_{a,b}|}\sum_{i\in\mathcal W_{a,b}}u_i^{(r,h)}v_i^{(r,h)}.
\end{equation}
The product requires both correct decisions on the same image. Marginal R4 acceptance and R3 rejection can remain high when their successes occur on different images.

For the IR margins \(m_{g,i}\) and \(m_{c,i}\), define the center \(b_i=(m_{g,i}+m_{c,i})/2\) and gap \(d_i=m_{g,i}-m_{c,i}\). Joint recovery at threshold \(\tau\) requires
\begin{equation}\label{eq:joint-threshold}
 m_{g,i}>\tau\ge m_{c,i}
 \quad\Longleftrightarrow\quad
 b_i-d_i/2 \le \tau < b_i+d_i/2.
\end{equation}
Same-image ordering counts \(d_i>0\). Pooled R4-versus-R3 AUC compares true and Base-error scores across images, with half credit for ties. Global AUC uses all primary true and false premises. Repair and damage count transitions from incorrect to correct and from correct to incorrect on shared valid primary questions.

\subsection{Primary endpoint and uncertainty}
The primary contrast is \(J_{43}^{(\GR)}(E)-J_{43}^{(\GR)}(A)\) within each dataset, model, and training seed. FAR accompanies the primary contrast, since a verifier can gain joint successes while accepting more false claims elsewhere. We also report CountAcc, TAR, BA, fixed-reference TCFA, and the corresponding default IR metrics. Per-arm metrics use each arm's valid responses. Paired verification contrasts use the shared valid conditions. Each interval uses 2,000 paired image-cluster bootstrap draws using a fixed random seed. Seed-level effects are summarized with equal weights and their standard deviation. Appendix~\ref{app:dev-calibration} describes the operating-point analysis on Dev.

\begin{figure}[!tbp]
\centering
\includegraphics[width=\textwidth]{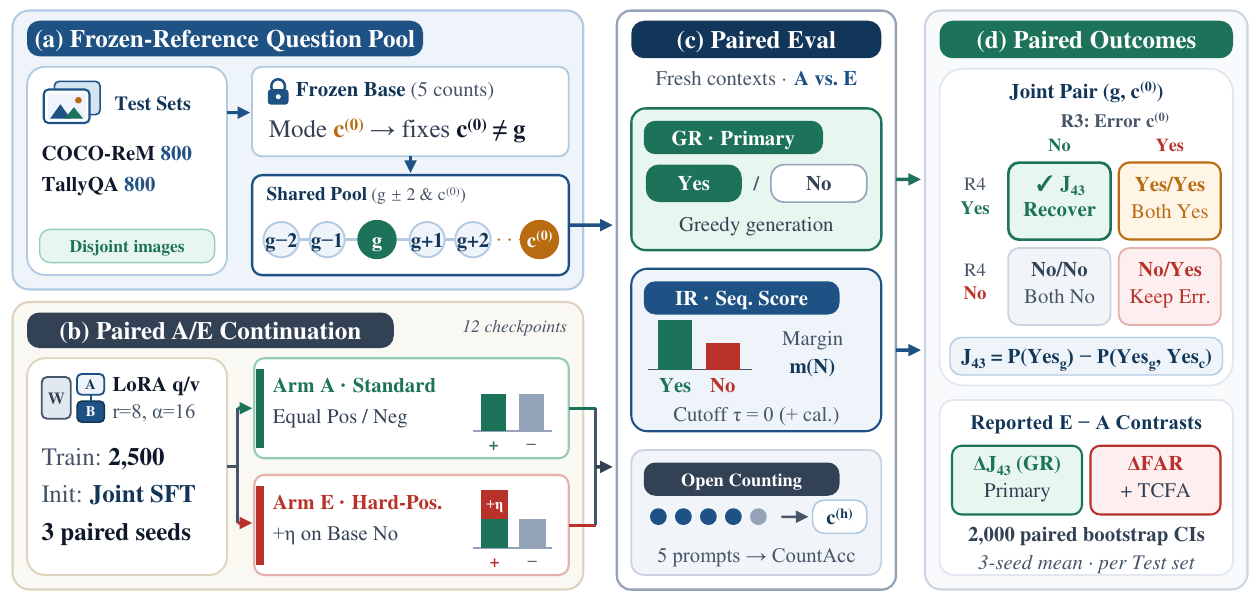}
\caption{Matched A/E training and independent evaluation protocol. \textbf{(a)}~Frozen Base open counts define a fixed per-model error cohort (\(c^{(0)}\ne g\)) and a shared verification pool on COCO-ReM Test and TallyQA Test. \textbf{(b)}~Standard continuation (A) and hard-positive continuation (E) branch from the same Joint SFT checkpoint with matched \(q/v\) LoRA adapters and paired seed order. \textbf{(c)}~Frozen adapters are evaluated in fresh contexts under generated (GR) and sequence-scored (IR) readouts alongside adapted open counting. \textbf{(d)}~Joint recovery \(J_{43}\) isolates the \Yes/\No{} cell on \((g,c^{(0)})\) and is reported alongside companion FAR with paired image-cluster bootstrap intervals.}
\Description{A four-panel pipeline diagram of the paired training and independent evaluation protocol. Panel a shows COCO-ReM Test and TallyQA Test feeding into Frozen Base open counting to fix c0 and create a shared verification pool combining the ground-truth window and the Base error. Panel b shows COCO-ReM Train and a shared Joint SFT initialization splitting into matched LoRA q/v Arms A and E across three paired seeds. Panel c shows both tracks converging into paired evaluation under GR, IR, and open counting. Panel d shows the 2 by 2 joint decision matrix defining J43 and the primary E minus A contrasts with paired bootstrap intervals.}
\label{fig:eval-flow}
\end{figure}

\subsection{Training and independent evaluation}
Each model's shared Joint SFT initialization trains a \(q/v\) LoRA adapter~\cite{hu2022lora} (\(r=8\), \(\alpha=16\), dropout 0) for three epochs on the 2,500 COCO-ReM Train labels (7,500 updates, AdamW~\cite{loshchilov2019adamw} at learning rate \(10^{-4}\)) using counting cross-entropy plus balanced true and false verification cross-entropy drawn from \(g\pm 2\). Continuation then runs 1,702 optimizer updates for Qwen3.5 and 1,772 for Gemma-3 at learning rate \(2\times10^{-5}\), weight decay 0.01, and gradient clipping at 1. A and E share training prompts and update order within each seed, and all three seeds of E use the same frozen-Base hard-positive subset on Train, so cross-seed variation reflects initialization and batch shuffling rather than subset drift. For Qwen3.5, E reuses the fixed A order for each seed; for Gemma seeds 1 and 2, grouped and counting-only batches are shuffled separately while preserving phase order. The final \(q/v\) Gemma adapter trains 2,228,224 parameters, compared with 3,203,072 for its \(q/k/v\) Dev variant.

Figure~\ref{fig:eval-flow} summarizes the paired training and independent evaluation protocol. We evaluate 12 fixed A/E checkpoints on both Test sets. Within each model and dataset, all arms and seeds share the same frozen Base open counts. On TallyQA, the original question serves as the first open-count prompt and the remaining four preserve its target expression. Prompts, parsers, splits, and scoring rules remain identical across both Test sets, while Dev operating-point comparisons stay on their five-fold splits. Split manifests, prompt templates, evaluation scripts, and per-image results are anonymously available at \url{https://anonymous.4open.science/r/tcfa-experiment-results-A424/}.

\section{Results}\label{sec:findings}
\subsection{Cross-family failures and the Joint SFT trade-off}
Balanced accuracy can conceal opposing verification profiles. Under GR, Qwen3.5 and Phi obtain 60.08\% and 60.09\% BA (Table~\ref{tab:base}). Their TAR is 25.80\% and 73.00\%, with FAR of 5.64\% and 52.82\%, respectively. Changes in the readout also move both rates within the same model. For LFM, IR lowers TCFA from 41.34\% to 14.18\% and TAR from 73.80\% to 53.40\%. BA changes from 65.90\% to 65.83\%.

\begin{table}[!tbp]
\caption{Six Base checkpoints from four families on COCO-ReM Dev (500 images, 2,450 shared primary verification conditions). Rates are percentages. CountAcc uses 500 images. Conditional TCFA uses the model's own defined open-count reference.}
\label{tab:base}
\centering\footnotesize\setlength{\tabcolsep}{3pt}
\begin{tabular}{@{}lcrrrrr@{}}
\toprule
Model & Readout & CountAcc $\uparrow$ & FAR $\downarrow$ & TAR $\uparrow$ & BA $\uparrow$ & TCFA $\downarrow$ \\
\midrule
\datainput evidence/base_rows.tex
\bottomrule
\end{tabular}
\tabnote{Defined modes in row-group order are 484, 496, 491, 478, 495, and 490. Qwen2.5 GR has seven invalid primary responses, including five true and two false premises. Gemma GR has 16 invalid false-premise responses, including five R5 candidates. Other rows have 2,450 valid primary decisions.}
\end{table}

\paragraph{Five-relation decision profiles}
Endorsing the open-count error and endorsing a third number vary across models. Table~\ref{tab:rbyfamily} shows both on the same ground-truth-centered candidate window. Under GR, Qwen3.5 accepts the erroneous open count in 18.36\% of R3 conditions and recognizes the true count in 5.18\% of R4 conditions. Phi obtains 88.46\% and 63.40\%, respectively. Third-count acceptance is 1.76\% for Qwen3.5 and 51.93\% for Phi. The acceptance rates use valid conditions in the shared ground-truth window. Group R3 measures endorsement of the open error, while R5 measures endorsement of an additional false quantity. Their rates differ across all six checkpoints in Table~\ref{tab:rbyfamily}.

\begin{table}[!tbp]
\caption{Five-relation Base decomposition on COCO-ReM Dev. Entries show \Yes{} rates in percent, followed by the number of valid conditions. R3 uses the primary window. Each model uses its own defined open-count mode, and R5 acceptance measures TCFA.}
\label{tab:rbyfamily}
\centering\footnotesize\setlength{\tabcolsep}{6pt}
\begin{tabular}{@{}lcrrrrr@{}}
\toprule
Model & Readout & R1 & R2 & R3 & R4 & R5 \\
\midrule
\datainput evidence/relation_rows.tex
\bottomrule
\end{tabular}
\end{table}

\paragraph{Joint SFT}\label{sec:sft-finding}
Joint SFT improves open counting while increasing false approval on a fixed set of premises. Table~\ref{tab:sft-finding} compares Qwen3.5 Base and ordinary Joint SFT on the same 500 images and 2,450 primary verification questions. Joint SFT increases the number of images counted correctly from 182 to 226. Under generated answers, TAR rises by 25.20 percentage points and FAR by 11.74 points. Balanced accuracy increases by 6.73 points because the true-premise gain is larger on the class-normalized scale. Sequence scoring shows the same direction of change in TAR, FAR, and balanced accuracy.

Across the fixed questions, SFT accepts 126 more true premises and 229 more false premises in net terms, about 1.82 additional false acceptances per additional true acceptance. At the benchmark's mix of 500 true and 1,950 false premises, this amounts to 103 fewer correct verification decisions. Class balancing helps compare systems independent of that mix. The raw counts describe the impact at the tested workload. A caption auditor could retain more accurate descriptions while also passing more incorrect counts for publication.

Consistency-based supervision teaches related question formats together~\cite{quantity}. Its value depends on whether the responses agree with the image. A verifier can agree with an erroneous open count, while correcting an error requires accepting a different number. Joint SFT is our example of shared supervision for the two interfaces. The observed co-occurrence of counting gains and increased false acceptance motivates evaluating the direction and correctness of agreement alongside aggregate scores.

Model-specific TCFA also rises from 18/1,022 to 92/858 under GR. Fine-tuning changes which images the model miscounts, so the model-specific denominators move. The increase in FAR on the same 1,950 false prompts establishes the verification trade-off on a common workload. These results motivate comparing targeted supervision at a matched false-acceptance budget.

\begin{table}[!tbp]
\caption{Qwen3.5 Base and Joint SFT on COCO-ReM Dev. Counting accuracy uses all 500 images. TAR and FAR use the same 500 true and 1,950 false primary questions. Entries are percentages.}
\label{tab:sft-finding}
\centering\small
\begin{tabular}{@{}llrrrrr@{}}
\toprule
Model & Readout & CountAcc & TAR & FAR & BA & Own TCFA\\
\midrule
\datainput evidence/sft_rows.tex
\bottomrule
\end{tabular}
\tabnote{TCFA uses the corresponding model's own eligible questions. Counts in row order are 18/1,022, 92/858, 40/1,022, and 78/858.}
\end{table}

\subsection{Score centers, gaps, and threshold calibration}\label{sec:recovery}
Base GR false acceptance varies widely across models (5.64\% for Qwen3.5, 48.29\% for Gemma-3, and 52.82\% for Phi-3.5). The E rule, which selects true premises rejected by frozen Base, covers 1,552 of the 1,659 Qwen R4 positives (93.55\%) and 746 of the 1,654 Gemma R4 positives (45.10\%). The same training rule therefore supplies different supervision relative to the Base counting-error cohort.

In the wider Gemma adapter setting, targeted continuation shifts the scores toward joint acceptance. In the Gemma \(q/k/v\) pilot, D and E raise both IR scores on 300 and 282 of 318 fixed Base-error images, respectively. The median center shifts are \(+0.5625\) and \(+0.3750\), and the signed median gap shifts round to zero. Median absolute center shifts are 0.5625 and 0.3750, compared with median absolute gap shifts of 0.2500 and 0.1251. At the default cutoff, double acceptance rises from 222 images under A to 279 under D and 265 under E. Equation~\ref{eq:joint-threshold} explains the decision change. Raising both margins can bring both above the cutoff while keeping their relative order.

Gemma \(q/v\) shows median center and gap shifts near zero, with default IR double acceptances increasing from 243 under A to 256 under D and 257 under E. The truth strictly outranks the Base error on 159, 162, and 166 images, respectively. The \(q/k/v\) counts are 160, 165, and 168 (Table~\ref{tab:ordering}). The pooled AUC can lie below 0.5 while most within-image pairs have the correct order, because it compares scores across images. The center-shift pattern is clearest with \(q/k/v\). D and E start from each model's Joint SFT checkpoint, while their selection rules use frozen Base outputs. Base acceptance and selection overlap describe the antecedents to training. The paired score changes characterize how each continuation changes decisions.

\begin{table}[!tbp]
\centering\small
\caption{Threshold-independent IR score summaries on fixed Base-error images, seed 0. Pooled AUC compares all R4 scores with all R3 scores. Same-image order counts \(m_g>m_c\). Center and gap shifts are medians of within-image changes from A.}
\label{tab:ordering}
\begin{tabular}{@{}lcrrrrr@{}}
\toprule
Model & Arm & R4 vs.\ R3 AUC & Same-image order & Ties & Center shift & Gap shift\\
\midrule
\datainput evidence/ranking_rows.tex
Gemma-3 $q/v$ & A & 0.4544 & 159/318 & 0 & +0.0000 & +0.0000 \\
Gemma-3 $q/v$ & D & 0.4589 & 162/318 & 0 & +0.0000 & +0.0000 \\
Gemma-3 $q/v$ & E & 0.4634 & 166/318 & 0 & +0.0000 & +0.0000 \\
\bottomrule
\end{tabular}
\end{table}

\begin{table}[!tbp]
\centering\small
\caption{Joint IR states on fixed Base-error images, seed 0. Each pair lists the response to the annotation first and to the Base error second. Yes/No is successful recovery. The calibrated columns use the 20\% FAR budget.}
\label{tab:states}
\begin{tabular}{@{}lc*{8}{r}@{}}
\toprule
& & \multicolumn{4}{c}{Default IR} & \multicolumn{4}{c}{Out-of-fold calibrated IR}\\
\cmidrule(lr){3-6}\cmidrule(lr){7-10}
Model & Arm & No/No & No/Yes & Yes/No & Yes/Yes & No/No & No/Yes & Yes/No & Yes/Yes\\
\midrule
\datainput evidence/state_rows.tex
Gemma-3 $q/v$ & A & 11 & 38 & 26 & 243 & 136 & 76 & 58 & 48 \\
Gemma-3 $q/v$ & D & 9 & 32 & 21 & 256 & 137 & 72 & 59 & 50 \\
Gemma-3 $q/v$ & E & 9 & 33 & 19 & 257 & 141 & 75 & 60 & 42 \\
\bottomrule
\end{tabular}
\end{table}

\paragraph{Calibration as a decision baseline}\label{sec:cal-results}
Calibrating fixed scores is a strong first-line baseline that recovers substantially more Gemma pairs without retraining. Table~\ref{tab:calibration} keeps the A adapter and its IR scores fixed. For each of five image-grouped folds, we maximize joint recovery under the FAR budget on the other four folds and apply the cutoff to the held-out images. With the formal \(q/v\) adapter at the 20\% budget, calibration alone raises Gemma-3 A from 26 to 58 of 318 joint recoveries (\(+10.06\pp\), paired 95\% CI \([5.23, 15.11]\)) while cutting realized FAR from 65.95\% to 18.97\%---exceeding every uncalibrated training arm. The wider Gemma \(q/k/v\) adapter shows the same pattern (\(+28\) pairs, \(+8.81\pp\) \([3.57, 14.33]\), FAR \(60.26\%\to 19.69\%\)). For Qwen3.5 seed-0 A, whose default FAR already sits near 20\% (20.97\%), the 20\% cutoff leaves joint recovery virtually unchanged (115 to 114 at 19.33\% FAR), and tightening to 10\% lowers recovery across all three seeds.

\begin{figure}[!tbp]
\centering
\includegraphics[width=.97\textwidth]{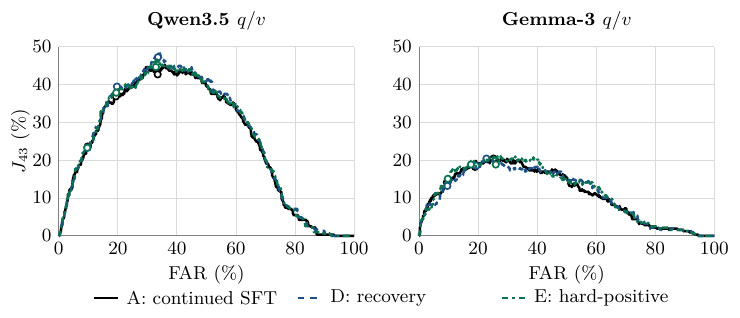}
\caption{FAR--\(J_{43}\) curves for seed 0 using the formal \(q/v\) adapters (Qwen3.5 \(q/v\) and Gemma-3 \(q/v\)). Lines sweep a common threshold on the full Dev set for each arm. Open markers show out-of-fold estimates at FAR budgets of 10\%, 20\%, and 40\%, positioned at their realized FAR.}
\Description{Two panels compare continuation, relation-conditioned training, and hard-positive training for Qwen q/v and Gemma q/v. Solid, dashed, and dash-dotted lines show the full threshold sweeps. Open markers show independently calibrated out-of-fold decisions.}
\label{fig:far-j43}
\end{figure}

Figure~\ref{fig:far-j43} shows the full-Dev threshold scans for the formal \(q/v\) adapters and marks out-of-fold performance at each budget. For a correctly ordered pair, lowering the cutoff below the truth score adds a joint success, whereas lowering it below the Base-error score converts that success into a double acceptance. Different pairs cross these boundaries at different cutoffs, so joint recovery peaks while FAR continues to rise.

\begin{table}[!tbp]
\centering\small
\caption{Effect of threshold calibration alone on arm A, seed 0. The joint pair denominator is held fixed. FAR values are percentages. The last column gives the change in \(J_{43}\) in percentage points with paired 95\% intervals.}
\label{tab:calibration}
\begin{tabular}{@{}lrrrrrl@{}}
\toprule
Model & Budget (\%) & \multicolumn{2}{c}{Joint successes} & \multicolumn{2}{c}{Actual FAR} & \(\Delta J_{43}\) [95\% CI]\\
\cmidrule(lr){3-4}\cmidrule(lr){5-6}
& & Default & Calibrated & Default & Calibrated &\\
\midrule
\datainput evidence/calibration_rows.tex
Gemma-3 $q/v$ & 20 & 26/318 & 58/318 & 65.95 & 18.97 & +10.06 [5.23, 15.11] \\
\bottomrule
\end{tabular}
\end{table}

\paragraph{Targeted continuation under FAR budgets}
At the 20\% FAR budget, Qwen3.5 D improves \(J_{43}\) over calibrated A across three seeds by 2.59, 0.97, and 0.97 percentage points (Table~\ref{tab:budgets}). At 40\%, the larger gain at seed 0 gives way to small declines at seeds 1 and 2. For Gemma-3 \(q/k/v\), D trails A at all three budgets. E improves \(J_{43}\) at the 40\% budget from 59 to 70 of 318 pairs at a realized FAR of 26.97\%, compared with 29.13\% for A. The paired results vary by objective and operating point. Appendix~\ref{sec:budget-results} gives the detailed decomposition.

\begin{table}[!tbp]
\centering\footnotesize\setlength{\tabcolsep}{2.5pt}
\caption{Out-of-fold A/D/E comparison under fixed FAR budgets. FAR caps are percentages. A, D, and E show joint successes divided by valid Base-error pairs, followed by actual FAR in percent. Effect columns report percentage-point differences with paired 95\% intervals. The Gemma \(q/k/v\) experiment is the adapter-scope sensitivity analysis.}
\label{tab:budgets}
\begin{tabular}{@{}lcrrrrrl@{}}
\toprule
Model & Seed & FAR cap & A & D & E & D--A [95\% CI] & E--A [95\% CI]\\
\midrule
\datainput evidence/budget_rows.tex
\bottomrule
\end{tabular}
\end{table}

\paragraph{Development evidence for the A/E comparison}
With aligned \(q/v\) adapters, Gemma seed 0 moves from 74 to 77 joint recoveries out of 318 under default GR when replacing A with E. FAR rises from 23.64\% to 38.67\%. The strictly ordered IR pairs increase from 159 to 166 of 318. Calibrating A alone at a 20\% FAR budget raises IR joint recovery from 26 to 58 pairs. Under the same budget, E recovers 60 pairs at 17.59\% realized FAR, compared with 18.97\% for A. At the 40\% budget, E changes joint recovery relative to A by \(-1.57\) percentage points with \(q/v\) and \(+3.46\) with \(q/k/v\). The Dev sensitivity analysis in Appendix~\ref{app:dev-calibration} also includes an offline score diagnostic showing that A and E yield the same fraction of truth-centered local peaks on both \(q/v\) models (Appendix~\ref{app:local-peak}).

\FloatBarrier
\subsection{Independent evaluation of targeted continuation}\label{sec:test}
COCO-ReM Test measures transfer to unseen images from the same annotation source, whereas TallyQA Test evaluates external simple category-count questions on Visual Genome images. Table~\ref{tab:independent-ae} summarizes three-seed equal-weight means and standard deviations on shared valid conditions, and Table~\ref{tab:coco-rem-seeds} reports the per-seed paired recovery contrasts on both benchmarks. The fixed Base-error cohorts contain 516 Qwen3.5 images and 533 Gemma-3 images on COCO-ReM Test, and 189 Qwen3.5 images and 370 Gemma-3 images on TallyQA Test. The two benchmarks also differ sharply in error magnitude: on TallyQA Test, 75.1\% of Qwen3.5 Base errors (142/189) and 72.7\% of Gemma-3 errors (269/370) are off by one (\(|c^{(0)}-g|=1\)), compared with 43.4\% (224/516) and 45.6\% (243/533) on COCO-ReM Test (where 33.1\% and 31.9\% differ by three or more).

Across all 24 paired comparisons (two datasets, two models, three seeds, and two readouts), hard-positive continuation (E) raises TAR, FAR, and fixed-reference TCFA in every single run, with all 72 image-level 95\% paired bootstrap intervals strictly above zero. Whether higher truth acceptance translates into paired same-image recovery \(J_{43}\) depends on both model family and dataset.

\paragraph{Qwen3.5: in-source recovery gains do not transfer to TallyQA}
On COCO-ReM Test, E raises default GR \(J_{43}\) from 41.47\% to 45.09\% (\(+3.62\pm 1.86\pp\)), with positive point estimates in all three seeds (seed~2 lower CI bound 0.00) and a \(+2.91\pp\) lift under default IR (Table~\ref{tab:coco-rem-seeds}). Under GR, global TAR rises from 57.42\% to 62.54\%, FAR rises from 18.42\% to 22.48\%, and fixed-reference TCFA rises from 20.69\% to 25.70\% (\(+5.02\pp\)). On external TallyQA Test, the recovery gain disappears: GR \(J_{43}\) shifts from 27.34\% to 26.81\% (\(-0.53\pm 1.40\pp\)), with per-seed changes of \(+1.06\), \(-1.59\), and \(-1.06\pp\) whose 95\% intervals all span zero, and IR \(J_{43}\) shifts by \(-0.88\pp\). Meanwhile, GR TAR rises from 70.79\% to 72.67\%, FAR climbs from 12.25\% to 13.93\% (\(+1.68\pp\)), and fixed-reference TCFA climbs from 14.62\% to 17.27\% (\(+2.66\pp\)).

\paragraph{Gemma-3: double acceptance erodes paired recovery on both benchmarks}
On COCO-ReM Test, E pushes global GR TAR from 58.92\% to 81.79\%, while FAR jumps from 30.87\% to 57.03\% and fixed-reference TCFA jumps from 27.35\% to 56.57\% (\(+29.22\pp\)). Mean \(J_{43}\) drops by \(5.69\pp\) under GR and \(4.07\pp\) under IR. At seed~2 under GR, E approves the annotation on 455 of 533 Base-error images (up from 239 under A), yet simultaneous \Yes/\Yes{} approvals jump from 112 to 382, cutting joint recovery from 127 to 73. TallyQA Test reproduces this failure even more sharply: GR \(J_{43}\) falls by half from 29.10\% to 14.59\% (\(-14.50\pm 7.99\pp\)), declining substantially in every seed (\(-7.57\), \(-12.70\), and \(-23.24\pp\)), while IR \(J_{43}\) drops by \(3.96\pp\). Across the 370 fixed Base-error images on TallyQA, GR truth acceptance rises by \(16.13\pp\) (69.19\% to 85.32\%), whereas simultaneous \Yes/\Yes{} acceptance jumps by \(30.63\pp\) (40.09\% to 70.72\%). Global GR FAR increases from 24.23\% to 39.70\% (\(+15.47\pp\)), and fixed-reference TCFA rises from 23.37\% to 41.70\% (\(+18.33\pp\)).

Across the three seeds, open-count accuracy barely moves on either benchmark (COCO-ReM: Qwen3.5 45.96\% to 45.88\%, Gemma-3 37.88\% to 37.00\%; TallyQA: Qwen3.5 69.75\% to 70.12\%, Gemma-3 60.83\% to 60.75\%). Mean GR balanced accuracy also obscures the collapse: on COCO-ReM it moves from 69.50\% to 70.03\% for Qwen3.5 and from 64.03\% to 62.38\% for Gemma-3, and on TallyQA from 79.27\% to 79.37\% for Qwen3.5 and from 77.43\% to 75.65\% for Gemma-3---changing by less than two points even as Gemma-3's paired recovery drops by half.

\paragraph{Locked Dev-to-Test calibration transfer and Test score geometry}
To test whether E's recovery differences persist at matched false-acceptance risk, we lock the Dev-fitted 20\% FAR IR thresholds for seed~0 (\(\tau_A=0.1142, \tau_E=0.2401\) for Qwen3.5; \(\tau_A=\tau_E=1.5000\) for Gemma-3 \(q/v\)) and apply them unchanged to both Test sets (Table~\ref{tab:test-transfer}). Two results stand out. First, once realized FAR is aligned on COCO-ReM Test (18.01\% vs.\ 18.37\%), Qwen3.5 E's default \(+3.10\pp\) IR advantage vanishes completely (\(-0.19\pp\), CI \([-1.74, 1.16]\)), and Gemma-3 A and E tie at 74/533 (13.88\%, FAR 19.07\% vs.\ 17.88\%). Second, on TallyQA Test, Gemma-3's locked Dev threshold shifts realized FAR slightly above the 20\% target (23.77\% for A, 22.01\% for E), yet calibration alone raises paired recovery by \(+21.08\pp\) on A (\(25\to 103\) of 370, CI \([16.76, 25.68]\)) and \(+25.14\pp\) on E (\(18\to 111\) of 370, CI \([20.54, 30.27]\)), whereas E minus A remains statistically indistinguishable from zero (\(+2.16\pp\) \([-0.54, 4.86]\)). Across all three seeds on both Test sets, IR same-image pair ordering (\(m_g > m_c\)) is nearly identical for A and E (COCO-ReM: Qwen3.5 72.61\% vs.\ 73.00\%, Gemma-3 48.16\% vs.\ 48.34\%; TallyQA: Qwen3.5 40.04\% vs.\ 39.68\%, Gemma-3 62.97\% vs.\ 63.69\%), with median gap shifts within \([-0.125, +0.007]\) while median pair centers rise across seeds (up to \(+0.190\) for Qwen3.5 and \(+0.688\) for Gemma-3). Thus, E shifts default operating points rather than improving within-image discrimination on either Test set.

\paragraph{Residual third-count acceptance on recovered images}
Because R5 premises are a subset of false premises, rising TCFA alongside FAR alone does not prove that TCFA adds information beyond \(J_{43}\) and FAR. We therefore measure \(\mathrm{R5}\mid J_{43}\): the fraction of images passing \(J_{43}\) that still endorse at least one R5 third-count candidate in the \(g\pm 2\) window. Under default GR (three-seed means), residual third-count acceptance is widespread even on successfully recovered images: 53.82\%\(\to\)62.71\% for Qwen3.5 (e.g., seed~0: 125/223\(\to\)155/240) and 71.10\%\(\to\)86.79\% for Gemma-3 (seed~0: 76/119\(\to\)103/126) on COCO-ReM Test, and 43.20\%\(\to\)53.37\% for Qwen3.5 and 50.37\%\(\to\)53.05\% for Gemma-3 on TallyQA Test. Even after Dev-locked 20\% FAR calibration (Table~\ref{tab:test-transfer}), 36.0\%--62.2\% of recovered images still endorse a third wrong count (and on calibrated COCO-ReM Gemma-3, E rises from 40/74 [54.1\%] to 46/74 [62.2\%] despite a lower overall FAR). Paired recovery alone therefore cannot certify that a verifier rejects incompatible counts.

\begin{table}[!tbp]
\centering\footnotesize
\setlength{\tabcolsep}{2.6pt}
\caption{Independent A/E evaluation on COCO-ReM Test and external TallyQA Test (800 images each). Rates are percentages averaged with equal weight across three seeds; \(\Delta\) columns report \(E-A\) in percentage points (\(\text{mean}\pm\text{SD}\) across seeds). GR is primary and IR is default sequence-scored.}
\label{tab:independent-ae}
\begin{tabular}{@{}lllrrrrrrr@{}}
\toprule
 & & & \multicolumn{3}{c}{\(J_{43}\)} & \multicolumn{3}{c}{FAR} & \multicolumn{1}{c}{TCFA}\\
\cmidrule(lr){4-6}\cmidrule(lr){7-9}\cmidrule(lr){10-10}
Dataset & Model & Readout & A & E & \(\Delta\ (\text{mean}\pm\text{SD})\) & A & E & \(\Delta\ (\text{mean}\pm\text{SD})\) & \(\Delta\)\\
\midrule
COCO-ReM & Qwen3.5 & GR & 41.47 & 45.09 & \(+3.62\pm 1.86\) & 18.42 & 22.48 & \(+4.06\pm 0.61\) & +5.02 \\
 &  & IR & 41.34 & 44.25 & \(+2.91\pm 0.51\) & 18.15 & 21.79 & \(+3.64\pm 0.83\) & +4.76 \\
 & Gemma-3 & GR & 22.89 & 17.20 & \(-5.69\pm 6.14\) & 30.87 & 57.03 & \(+26.16\pm 12.88\) & +29.22 \\
 &  & IR & 12.20 & 8.13 & \(-4.07\pm 2.33\) & 64.57 & 75.63 & \(+11.06\pm 7.58\) & +12.76 \\
\midrule
TallyQA & Qwen3.5 & GR & 27.34 & 26.81 & \(-0.53\pm 1.40\) & 12.25 & 13.93 & \(+1.68\pm 0.44\) & +2.66 \\
 &  & IR & 26.81 & 25.93 & \(-0.88\pm 1.70\) & 12.41 & 13.73 & \(+1.32\pm 0.23\) & +2.02 \\
 & Gemma-3 & GR & 29.10 & 14.59 & \(-14.50\pm 7.99\) & 24.23 & 39.70 & \(+15.47\pm 7.94\) & +18.33 \\
 &  & IR & 8.74 & 4.77 & \(-3.96\pm 2.51\) & 46.30 & 54.19 & \(+7.89\pm 5.97\) & +7.86 \\
\bottomrule
\end{tabular}
\end{table}

\begin{table}[!tbp]
\centering\footnotesize
\setlength{\tabcolsep}{4.0pt}
\caption{Per-seed joint recovery on COCO-ReM Test and TallyQA Test. Counts give A to E successes on shared Base-error cohorts (COCO-ReM: 516 Qwen3.5, 533 Gemma-3; TallyQA: 189 Qwen3.5, 370 Gemma-3). Differences are in percentage points with paired 95\% image-bootstrap intervals.}
\label{tab:coco-rem-seeds}
\begin{tabular}{@{}llcrrlrrl@{}}
\toprule
 & & & \multicolumn{3}{c}{Default GR} & \multicolumn{3}{c}{Default IR}\\
\cmidrule(lr){4-6}\cmidrule(lr){7-9}
Dataset & Model & Seed & A\(\to\)E & \(\Delta J_{43}\) & [95\% CI] & A\(\to\)E & \(\Delta J_{43}\) & [95\% CI]\\
\midrule
COCO-ReM & Qwen3.5 & 0 & 223\(\to\)240 & +3.29 & [1.71, 5.10] & 221\(\to\)237 & +3.10 & [1.40, 4.84] \\
 &  & 1 & 202\(\to\)231 & +5.62 & [3.04, 8.19] & 206\(\to\)218 & +2.33 & [0.38, 4.25] \\
 &  & 2 & 217\(\to\)227 & +1.94 & [0.00, 3.90] & 213\(\to\)230 & +3.29 & [1.24, 5.28] \\
 & Gemma-3 & 0 & 119\(\to\)126 & +1.31 & [-2.26, 4.72] & 63\(\to\)49 & -2.63 & [-4.30, -1.14] \\
 &  & 1 & 120\(\to\)76 & -8.26 & [-11.73, -4.87] & 57\(\to\)42 & -2.81 & [-4.85, -0.92] \\
 &  & 2 & 127\(\to\)73 & -10.13 & [-14.37, -6.05] & 75\(\to\)39 & -6.75 & [-9.61, -4.13] \\
\midrule
TallyQA & Qwen3.5 & 0 & 52\(\to\)54 & +1.06 & [-0.65, 3.21] & 50\(\to\)52 & +1.06 & [-1.20, 3.41] \\
 &  & 1 & 51\(\to\)48 & -1.59 & [-5.61, 2.20] & 50\(\to\)47 & -1.59 & [-4.62, 1.04] \\
 &  & 2 & 52\(\to\)50 & -1.06 & [-3.70, 1.48] & 52\(\to\)48 & -2.12 & [-5.18, 0.56] \\
 & Gemma-3 & 0 & 106\(\to\)78 & -7.57 & [-11.31, -3.85] & 25\(\to\)18 & -1.89 & [-3.44, -0.77] \\
 &  & 1 & 92\(\to\)45 & -12.70 & [-16.46, -8.97] & 31\(\to\)19 & -3.24 & [-5.11, -1.55] \\
 &  & 2 & 125\(\to\)39 & -23.24 & [-28.25, -18.45] & 41\(\to\)16 & -6.76 & [-9.57, -4.20] \\
\bottomrule
\end{tabular}
\end{table}

\begin{table}[!tbp]
\centering\scriptsize
\setlength{\tabcolsep}{2.8pt}
\caption{Locked Dev-to-Test 20\% FAR IR calibration transfer (seed~0) and residual third-count acceptance on recovered images (\(\mathrm{R5}\mid J_{43}\): fraction of \(J_{43}\) successes endorsing \(\ge 1\) R5 candidate in \(g\pm 2\)). Thresholds fitted on COCO-ReM Dev are applied unchanged to both Test sets.}
\label{tab:test-transfer}
\begin{tabular}{@{}lllcccl@{}}
\toprule
Dataset & Model & Cutoff & FAR (A\(\to\)E) & \(J_{43}\) A\(\to\)E & \(\Delta J_{43}\) [95\% CI] & \(\mathrm{R5}\mid J_{43}\) (A\(\to\)E)\\
\midrule
COCO-ReM & Qwen3.5 & \(\tau=0\) & 18.75\(\to\)21.57 & 42.83\(\to\)45.93 & +3.10 [1.55, 4.84] & 123/221\(\to\)148/237 (55.7\(\to\)62.4\%) \\
 &  & Dev 20\% & 18.01\(\to\)18.37 & 42.25\(\to\)42.05 & -0.19 [-1.74, 1.16] & 118/218\(\to\)120/217 (54.1\(\to\)55.3\%) \\
 & Gemma-3 & \(\tau=0\) & 68.59\(\to\)73.49 & 11.82\(\to\)9.19 & -2.63 [-4.13, -1.13] & 56/63\(\to\)43/49 (88.9\(\to\)87.8\%) \\
 &  & Dev 20\% & 19.07\(\to\)17.88 & 13.88\(\to\)13.88 & +0.00 [-2.06, 2.25] & 40/74\(\to\)46/74 (54.1\(\to\)62.2\%) \\
\midrule
TallyQA & Qwen3.5 & \(\tau=0\) & 13.15\(\to\)14.35 & 26.46\(\to\)27.51 & +1.06 [-1.06, 3.70] & 22/50\(\to\)25/52 (44.0\(\to\)48.1\%) \\
 &  & Dev 20\% & 12.69\(\to\)12.60 & 25.93\(\to\)26.46 & +0.53 [-2.12, 3.17] & 21/49\(\to\)18/50 (42.9\(\to\)36.0\%) \\
 & Gemma-3 & \(\tau=0\) & 50.03\(\to\)52.82 & 6.76\(\to\)4.86 & -1.89 [-3.51, -0.54] & 11/25\(\to\)9/18 (44.0\(\to\)50.0\%) \\
 &  & Dev 20\% & 23.77\(\to\)22.01 & 27.84\(\to\)30.00 & +2.16 [-0.54, 4.86] & 44/103\(\to\)43/111 (42.7\(\to\)38.7\%) \\
\bottomrule
\end{tabular}
\tabnote{Corresponding fixed-reference TCFA (A\(\to\)E, \%) drops under Dev 20\% calibration from 20.73\(\to\)24.36 to 19.91\(\to\)20.14 (COCO-ReM Qwen3.5), 68.62\(\to\)74.13 to 14.29\(\to\)14.06 (COCO-ReM Gemma-3), 15.42\(\to\)17.68 to 14.56\(\to\)14.56 (TallyQA Qwen3.5), and 52.76\(\to\)55.60 to 21.80\(\to\)19.57 (TallyQA Gemma-3).}
\end{table}

\subsection{Fine-grained diagnostic slices}\label{sec:slices}
We split the existing Dev questions by true count and numerical discrepancy. In R5, discrepancy is \(|N-g|\). For a recovery pair, it is \(|c^{(0)}-g|\), because the other question is always the true premise. The paired slice includes a third bin for Base errors of three or more.

TCFA persists in the smaller-count stratum at a two-instance discrepancy. Phi-3.5 accepts 87 of 278 such premises, Gemma-3 accepts 113 of 260, and LFM2.5 accepts 56 of 227 (Table~\ref{tab:r5-slices}). Third-count acceptance therefore arises even among clearly separated, low-count alternatives; separating visual ambiguity from counting priors is left to future work.

The Gemma \(q/v\) E pilot yields opposite recovery changes in the two count bands. Under GR, the E--A contrast sums to \(-11\) successful pairs among 153 images with counts at most four and \(+14\) among 165 images with counts at least five (Table~\ref{tab:j43-slices}). The aggregate gain of three images combines changes in opposite directions.

\begin{table}[!tbp]
\centering\small
\caption{COCO-ReM Dev Base GR third-count acceptance by true count and \(|N-g|\). Entries are Yes / valid R5 conditions.}
\label{tab:r5-slices}
\begin{tabular}{@{}lrrrr@{}}
\toprule
 & \multicolumn{2}{c}{\(g=1\)--4} & \multicolumn{2}{c}{\(g=5\)--12}\\
\cmidrule(lr){2-3}\cmidrule(lr){4-5}
Model & Distance 1 & Distance 2 & Distance 1 & Distance 2\\
\midrule
Qwen3.5 & 4/211 & 3/250 & 6/280 & 5/281 \\
Phi-3.5 & 106/248 & 87/278 & 203/313 & 210/328 \\
Gemma-3 & 119/208 & 113/260 & 107/282 & 109/288 \\
LFM2.5 & 65/177 & 56/227 & 150/290 & 137/293 \\
\bottomrule
\end{tabular}
\end{table}

\begin{table}[!tbp]
\centering\footnotesize
\caption{COCO-ReM Dev GR joint recovery for \(q/v\), seed 0. Each cell shows A to E, with successful pairs divided by shared valid pairs. Grouping uses the discrepancy \(|c^{(0)}-g|\).}
\label{tab:j43-slices}
\begin{tabular}{@{}lrrrrrr@{}}
\toprule
 & \multicolumn{3}{c}{\(g=1\)--4} & \multicolumn{3}{c}{\(g=5\)--12}\\
\cmidrule(lr){2-4}\cmidrule(lr){5-7}
Model & 1 & 2 & \(\ge3\) & 1 & 2 & \(\ge3\)\\
\midrule
Qwen3.5 & \(43/85\to 44/85\) & \(16/39\to 18/39\) & \(11/24\to 13/24\) & \(13/42\to 13/42\) & \(15/41\to 16/41\) & \(19/78\to 24/78\) \\
Gemma-3 & \(23/96\to 14/96\) & \(10/30\to 8/30\) & \(20/27\to 20/27\) & \(6/48\to 12/48\) & \(7/42\to 10/42\) & \(8/75\to 13/75\) \\
\bottomrule
\end{tabular}
\end{table}

\FloatBarrier
\section{Discussion}\label{sec:discussion}
\subsection{Calibrate before retraining}
For a correctly ordered pair, the interval between the two scores gives the thresholds that yield joint success. Threshold calibration therefore serves as a high-leverage, zero-training baseline. On COCO-ReM Dev at a 20\% FAR budget, calibration alone adds 32 paired recoveries on Gemma-3 \(q/v\) A (\(26\to 58\) of 318) and 28 on \(q/k/v\), whereas Qwen3.5 A changes by one pair because its default FAR already lies near 20\%. Locking those same Dev thresholds and transferring them to external TallyQA Test reproduces this advantage on unseen Visual Genome images, lifting Gemma-3 A by \(+21.08\pp\) (\(25\to 103\) of 370) while simultaneously cutting FAR in half, whereas targeted retraining adds no statistically clear gain over calibrated A on either Test set. Practitioners deploying sequence-scored verifiers should therefore calibrate operating thresholds to their risk budget before attempting recovery fine-tuning.

\subsection{Targeted supervision and default decisions}
The formal A/E contrast evaluates the full targeted objective (\(\Loss_E\) vs.\ \(\Loss_A\)), which combines equal five-relation weighting, positive protection, and hard-positive upweighting. On the shared valid cohort \(\mathcal{W}_{A,E}\), Eq.~\eqref{eq:j43} expands identically into \(J_{43}=P(\Yes_g)-P(\Yes_g,\Yes_c)\), where \(\Yes_g\) and \(\Yes_c\) denote acceptance in fresh verification queries for the annotation \(g\) and the frozen Base error \(c^{(0)}\) (empirical frequencies on \(\mathcal{W}_{A,E}\), not statistical independence). Under default GR, truth acceptance within the Base-error cohort rises in all four model--dataset combinations: \(+6.07\pp\) for Qwen3.5 and \(+28.27\pp\) for Gemma-3 on COCO-ReM, and \(+3.70\pp\) for Qwen3.5 and \(+16.13\pp\) for Gemma-3 on TallyQA. Yet simultaneous \Yes/\Yes{} acceptance rises by \(+2.45\), \(+33.96\), \(+4.23\), and \(+30.63\pp\), respectively---outpacing the truth-acceptance gain in three of the four settings and leaving positive point estimates in all three seeds only for Qwen3.5 on COCO-ReM (\(+3.62\pp\), with seed~2's lower CI bound at 0.00).

Test-set IR score geometry confirms why: across all three seeds on both Test sets, E leaves same-image ordering (\(m_g>m_c\)) nearly flat (\(<0.8\pp\) shift) and never widens the median gap, while shifting pair centers upward. Consequently, matching FAR through locked Dev thresholds erases Qwen3.5's COCO-ReM advantage (\(-0.19\pp\)). On TallyQA, where over 72\% of Base errors are off-by-one (\(|c^{(0)}-g|=1\)), truth and error margins sit closer together, making simultaneous \Yes/\Yes{} transitions even more frequent under default thresholds. Because the two Base models reject very different subsets of true premises on Train, the same hard-positive rule also concentrates extra weight differently across families, while Dev score diagnostics confirm that A and E maintain identical local-peak totals (Appendix~\ref{app:local-peak}, where paired transitions are symmetric at 8 gained / 8 lost for Qwen3.5 and 17 / 17 for Gemma-3 \(q/v\)).

\subsection{Truth recovery and third-count acceptance}
The recovery pair tests \(g\) against the specific Base error \(c^{(0)}\), whereas TCFA tests another wrong number in \(g\pm 2\). Even among images that pass \(J_{43}\), 43\%--87\% under default GR and 36\%--62\% after locked 20\% FAR calibration still endorse at least one third wrong count (Table~\ref{tab:test-transfer}). On calibrated COCO-ReM Gemma-3, E raises this residual rate from 54.1\% to 62.2\% even at a lower overall FAR. A verifier can therefore accept three, reject four, and still endorse five on the same image; balanced accuracy averages separate pools and obscures both which false count was endorsed and whether the competing decisions were mutually consistent.

\subsection{Implications for multimedia evaluation}
A caption auditor and a numerical retrieval filter can use the same verifier with different tolerances for false approvals. A system at a strict budget may reject supported counts that it would retain in a human-reviewed workflow. Reporting the joint recovery rate at the attained FAR places that choice in operational terms. The paired test is particularly relevant when inventory review must approve one corrected record and reject the previous count using the same image. A separate TCFA measure records the exposure to other erroneous quantities, including ones the counting interface never proposed.

\subsection{Limitations}
Our Base diagnosis covers six open-weight checkpoints in the 3B--4B class, and the dual-Test training comparison evaluates two 4B families (Qwen3.5-4B and Gemma-3-4B-IT) with opposing Base priors. Larger checkpoints, hosted APIs, multi-image sequences, and complex attribute-qualified counts may behave differently; in particular, our TallyQA subset scopes to simple Visual Genome category-count questions (counts 1--12), and the A/E comparison evaluates the full targeted objective rather than isolating hard-positive selection from relation weighting and protection on Test. Cross-seed variance in Gemma-3 reflects training stochasticity under a shared frozen-Base hard-positive subset. Conditional recovery \(J_{43}\) applies to images with defined erroneous Base counts, while mode coverage and overall TAR/FAR characterize the full benchmark. Finally, sequence-score calibration requires log-probability access, whereas generated decisions characterize black-box text interfaces.

\section{Conclusion}\label{sec:conclusion}
Five numerical relations separate wrong-count agreement, truth recovery, and third-count false acceptance in small-parameter VLMs, revealing stark cross-family differences hidden by balanced accuracy. Paired recovery \(J_{43}\) measures whether a verifier accepts the annotation while rejecting its frozen Base error on the same image. Threshold calibration proves a strong first-line baseline: on COCO-ReM Dev it lifts Gemma-3 \(q/v\) recovery from 26 to 58 of 318 images without retraining, and locked Dev thresholds transfer to external TallyQA Test with a \(+21.08\pp\) recovery gain on Gemma-3 A while eliminating Qwen3.5's apparent COCO-ReM E advantage at matched FAR. Across 24 paired dual-Test evaluations, hard-positive continuation uniformly raises TAR, FAR, and TCFA yet leaves same-image score ordering nearly unchanged and fails to improve paired recovery once false-acceptance risk is controlled. Even among recovered images, 36\%--62\% after calibration still endorse a third wrong count. Practical numerical verification therefore requires calibrating risk before retraining and reporting \(J_{43}\), FAR, and TCFA together.

\paragraph{Code and data availability}
Split manifests, prompt templates, evaluation scripts, and per-image experimental results are anonymously available at \url{https://anonymous.4open.science/r/tcfa-experiment-results-A424/}.

\FloatBarrier
\bibliographystyle{ACM-Reference-Format}
\bibliography{references}

\clearpage
\appendix
\section*{Online Appendix}

\section{Development operating-point analysis}\label{app:dev-calibration}
\subsection{Default readouts and legacy matching on Dev}\label{sec:protocol-I}
Protocol I reports short-answer GR, zero-cutoff IR, and two legacy IR matches. For each of five image-grouped folds, the other four select the lowest cutoff meeting the reference FAR, or the cutoff closest to reference R4 acceptance. R4 ties favor lower acceptance, then a higher cutoff. D--A uses same-seed A for both targets. D--C uses same-seed C for FAR and A\(_0\) for R4.

\subsection{Out-of-fold FAR calibration on Dev}\label{sec:protocol-II}
The development sensitivity analysis examines FAR budgets of 10\%, 20\%, and 40\%. Given five image-grouped folds, for each arm \(h\) and held-out fold \(f\), the threshold is selected on the other four folds:
\begin{equation}\label{eq:calibration}
 \widehat{\tau}_{h,f}(B)\in
 \underset{\tau\,:\,\widehat{\mathrm{FAR}}_{h,-f}(\tau)\le B}{\arg\max}
 \ \widehat J_{43,h,-f}(\tau).
\end{equation}
The candidates are the primary false-premise scores, both scores from each recovery pair, and zero. Ties favor fewer false acceptances, then the higher cutoff. Held-out scores strictly above the cutoff receive \Yes{}.

Comparisons use shared valid questions and pairs. We pool held-out decisions to estimate realized FAR, TAR, BA, fixed-reference TCFA, and \(J_{43}\). Realized FAR can exceed the fitting budget. The full-Dev scans describe the available operating points. We use 2,000 paired image-cluster bootstrap draws with fixed fitted thresholds, so the intervals capture image-sampling variation conditional on those thresholds.

\subsection{Relation-conditioned training beyond calibration}\label{sec:budget-results}
For Qwen3.5 D, only the seed-0 interval lies above zero at the 20\% budget. At 10\%, the D--A changes are \(-0.32\), \(-0.97\), and \(+0.97\) points across seeds. Qwen seed-0 E gains 0.97 points with an interval spanning zero (Table~\ref{tab:budgets}).

At the 40\% budget, Gemma \(q/k/v\) E adds 18 joint successes and loses seven. Eleven of the newly successful images previously accepted both quantities, and seven previously rejected both. At 20\%, Gemma \(q/k/v\) D raises BA from 64.45\% to 64.95\% and lowers fixed R5 acceptance from 15.82\% to 14.57\%, while joint successes fall from 62 to 56. Gemma \(q/v\) E gains two joint successes at 20\% and loses five at 40\%. The accompanying \texttt{dev-budget-companion.csv} reports out-of-fold TAR, BA, and fixed-reference TCFA.

\paragraph{What the two supervision rules select}
On Train, E selects 1,878 Qwen3.5 positives across 2,224 update occurrences and 866 Gemma-3 positives across 1,031 occurrences. The extra coefficients are 0.14919 and 0.32085, respectively. Section~\ref{sec:recovery} reports their overlap with the R4 positives.

\subsection{Local-peak structure around the ground truth}\label{app:local-peak}
We use the COCO-ReM Dev \IR{} scores in the primary \(g\pm2\) window to characterize local ordering around the annotated count. The truth forms a \emph{local peak} when \(m(g)>\max\{m(g-1),m(g+1)\}\), and it ranks \emph{first in the window} when its score strictly exceeds every other primary candidate. The 50 images with \(g=1\) have four candidates, and the remaining 450 have five. Supplementary Base errors outside the primary window are excluded from these summaries.

Across the four Base models, the annotated count forms a local peak on 38.4\%--46.6\% of Dev images (Table~\ref{tab:local-peak}). For every model, \(g-1\) outranks \(g\) more often than \(g+1\) does. The asymmetry is largest for Gemma-3, with 285 versus 90 images. Its median truth-minus-neighbor gaps are \(-0.500\) for \(g-1\) and \(+2.250\) for \(g+1\) (Table~\ref{tab:local-asym}).

With \(q/v\) adapters at seed 0, Qwen3.5 A and E each have 233 local peaks out of 500 (225 shared, 8 gained, 8 lost), and Gemma-3 A and E each have 234 (217 shared, 17 gained, 17 lost). In the fixed Base-error cohorts, local peaks change from 108 to 111 of 309 for Qwen3.5 (34.95\% to 35.92\%) and from 114 to 112 of 318 for Gemma-3 (35.85\% to 35.22\%). In these cohorts, the median within-image changes in \(m(g)-m(g+1)\) and \(m(g)-m(g+2)\) range from \(-0.004\) to \(0\) for Qwen and are approximately \(-0.125\) for Gemma.

Any rule that requires a strict local peak to accept a candidate has \(J_{43}\) at most the fraction of truth-centered local peaks in the same cohort. For Qwen3.5 A\(_0\) and E\(_0\), those ceilings are 34.95\% and 35.92\%. The corresponding out-of-fold \(J_{43}\) values at the 20\% FAR budget are 36.89\% and 37.86\% on the same 309 images (Table~\ref{tab:budgets}).

\begin{table}[!htbp]
\centering\footnotesize
\setlength{\tabcolsep}{3pt}
\caption{Offline Dev \IR{} local-peak diagnostics in the \(g\pm2\) window. Monotonic decay requires both sides to decrease strictly at each step away from \(g\) on the 450 five-candidate images. A and E use \(q/v\) adapters and seed 0.}
\label{tab:local-peak}
\begin{tabular}{@{}lrrr@{}}
\toprule
Model / Arm & Local peak & Window-first & Monotonic decay\\
\midrule
Qwen3.5 Base & 217/500 (43.4\%) & 192/500 (38.4\%) & 116/450 (25.8\%)\\
Phi-3.5 Base & 192/500 (38.4\%) & 142/500 (28.4\%) & 66/450 (14.7\%)\\
Gemma-3 Base & 193/500 (38.6\%) & 144/500 (28.8\%) & 73/450 (16.2\%)\\
LFM2.5 Base & 233/500 (46.6\%) & 202/500 (40.4\%) & 131/450 (29.1\%)\\
\addlinespace
Qwen3.5 A \(\to\) E & \(233 \to 233\)/500 & \(222 \to 220\)/500 & \(139 \to 146\)/450\\
Gemma-3 A \(\to\) E & \(234 \to 234\)/500 & \(201 \to 206\)/500 & \(132 \to 133\)/450\\
\bottomrule
\end{tabular}
\end{table}

\begin{table}[!htbp]
\centering\small
\caption{Median within-image margin differences \(m(g)-m(g+k)\) for the COCO-ReM Dev Base \IR{} scores. Positive values favor the ground truth. The \(g-2\) column uses 450 images. The other columns use 500.}
\label{tab:local-asym}
\begin{tabular}{@{}lrrrr@{}}
\toprule
Base model & \(g-2\) & \(g-1\) & \(g+1\) & \(g+2\)\\
\midrule
Qwen3.5 & \(+0.388\) & \(+0.098\) & \(+0.250\) & \(+0.502\)\\
Phi-3.5 & \(+0.062\) & \(-0.063\) & \(+0.328\) & \(+0.500\)\\
Gemma-3 & \(+0.000\) & \(-0.500\) & \(+2.250\) & \(+3.875\)\\
LFM2.5 & \(+0.875\) & \(+0.125\) & \(+1.000\) & \(+1.625\)\\
\bottomrule
\end{tabular}
\end{table}

\FloatBarrier
\section{Default and legacy matched comparisons}\label{app:legacy}

\subsection{Development training controls}
Relation balancing (B) assigns equal weight to the five verification groups. Positive-loss protection (C) adds the reference penalty:
\begin{align}
 \Loss_B &= L_{\mathrm{count}}+\tfrac15\sum_{k=1}^{5}L_k,\\
 \Loss_C &= \Loss_B+P.
\end{align}
The D--C contrast isolates the extra R4 weight. For Group DRO~\cite{sagawa2020groupdro}, we use
\(\Loss_{\mathrm{DRO}}=L_{\mathrm{count}}+\sum_{k=1}^{5}w_kL_k\).
Starting from uniform weights, we multiply each \(w_k\) by \(\exp(0.01L_k)\) using detached group losses and renormalize to sum to one.

Table~\ref{tab:five-relation-ablation} gives the default GR and IR results for A/B/C/D.

\begin{table}[!htbp]
\centering
\caption{Four-arm ablation on COCO-ReM Dev with Qwen3.5-4B. All values are percentages at the default decision rules. FAR, TAR, and BA use the shared 2,450 primary questions. TCFA uses each adapted model's defined open-count reference. The fixed-Base paired success rate \(J_{43}\) uses 309 images. B contributes the seed-0 comparison.}
\label{tab:five-relation-ablation}
\footnotesize
\setlength{\tabcolsep}{4pt}
\begin{adjustbox}{max width=\linewidth}
\begin{tabular}{@{}lccrrrrrrrrrr@{}}
\toprule
& & & \multicolumn{5}{c}{Short-answer generation (GR)} & \multicolumn{5}{c}{Sequence scoring (IR)} \\
\cmidrule(lr){4-8}\cmidrule(lr){9-13}
Arm & Seed & CountAcc & FAR & TAR & BA & TCFA & \(J_{43}\) & FAR & TAR & BA & TCFA & \(J_{43}\) \\
\midrule
A & 0 & 44.20 & 20.92 & 54.60 & 66.84 & 14.84 & 37.86 & 20.97 & 54.80 & 66.91 & 14.73 & 37.22 \\
 & 1 & 45.60 & 19.79 & 56.80 & 68.50 & 12.97 & 36.89 & 19.38 & 56.60 & 68.61 & 12.03 & 33.33 \\
 & 2 & 45.20 & 20.41 & 56.40 & 67.99 & 13.42 & 37.54 & 20.62 & 57.80 & 68.59 & 13.19 & 38.51 \\
\addlinespace
B & 0 & 44.60 & 16.00 & 48.80 & 66.40 & 8.76 & 34.30 & 16.62 & 50.40 & 66.89 & 9.79 & 34.63 \\
\addlinespace
C & 0 & 44.40 & 18.87 & 53.60 & 67.36 & 11.76 & 37.86 & 20.15 & 54.80 & 67.32 & 13.38 & 37.22 \\
 & 1 & 45.40 & 18.46 & 56.00 & 68.77 & 11.40 & 34.30 & 18.26 & 56.40 & 69.07 & 10.93 & 33.33 \\
 & 2 & 45.20 & 18.77 & 54.20 & 67.72 & 11.05 & 33.98 & 19.54 & 54.60 & 67.53 & 11.75 & 33.98 \\
\addlinespace
D & 0 & 44.00 & 26.00 & 59.20 & 66.60 & 21.17 & 41.75 & 24.46 & 57.80 & 66.67 & 18.99 & 39.81 \\
 & 1 & 45.60 & 24.05 & 62.20 & 69.07 & 18.22 & 38.83 & 22.62 & 61.40 & 69.39 & 15.98 & 39.16 \\
 & 2 & 45.80 & 24.97 & 60.80 & 67.91 & 18.88 & 40.13 & 22.77 & 59.60 & 68.42 & 15.80 & 38.19 \\
\bottomrule
\end{tabular}
\end{adjustbox}
\end{table}

\subsection{Full objective versus ordinary continued fine-tuning}\label{sec:d-a}
Table~\ref{tab:da-mean} reports four settings for the full objective (D) and ordinary continuation (A): default GR, default IR, IR at matched FAR, and IR at matched R4 acceptance. These Protocol I comparisons retain the default decision rules and the reference-arm operating points for historical comparison. On 309 frozen Base miscounts, the default GR gains in \(J_{43}\) are 12, 6, and 8 additional jointly correct images, respectively. The mean paired gain is 2.80 points (two-sided 95\% Student \(t\) interval across three seed differences, 0.35 to 5.26, 2 degrees of freedom). At matched IR FAR the mean gain is 1.94 points, with seed differences of $+2.27$, $+3.88$, and $-0.32$. Table~\ref{tab:da} gives the image-level intervals within each seed.

Under GR, D raises TAR by 4.80 points and FAR by 4.63. These changes largely cancel in BA, which rises by 0.08 points. Of the 481 GR errors made by Base, 371 are rejected true premises. Repair increases by 4.44 points (49.34\% to 53.78\% on 481 errors), and damage increases by 4.46 points (18.99\% to 23.45\% on 1,969 correct decisions). Appendix Table~\ref{tab:repair-detail} gives the seed breakdown.
Table~\ref{tab:da-mean} gives third-count acceptance using each adapted model's own open count. Fixing the reference to the Base count gives identical eligible questions for A and D. On those questions, third-count acceptance changes from 22.21\% to 22.63\% at matched FAR and from 22.21\% to 22.11\% at matched R4. The seed-0 continuous scores provide another view. From A to D, pooled IR AUC on 2,450 primary premises changes from 0.7720 to 0.7683, and the number of images on which the truth strictly outranks the Base error changes from 218 to 219 out of 309. These score summaries use different candidate populations from joint recovery.

\begin{table}[!tbp]
\centering\footnotesize
\setlength{\tabcolsep}{3pt}
\caption{Qwen3.5 development results averaged over three seeds. Entries show A to D in percent. R5 uses each adapted model's own open count and the primary candidate window.}
\label{tab:da-mean}
\begin{tabular}{@{}lrrr@{}}
\toprule
Readout & Own R5 $\downarrow$ & $J_{43}\uparrow$ & FAR $\downarrow$ \\
\midrule
Default GR & 13.74 $\to$ 19.42 & 37.43 $\to$ 40.24 & 20.38 $\to$ 25.01 \\
Default IR & 13.31 $\to$ 16.92 & 36.35 $\to$ 39.05 & 20.32 $\to$ 23.28 \\
IR matched FAR & 13.31 $\to$ 13.22 & 36.35 $\to$ 38.30 & 20.32 $\to$ 20.31 \\
IR matched R4 & 13.31 $\to$ 13.03 & 36.35 $\to$ 37.97 & 20.32 $\to$ 19.93 \\
\bottomrule
\end{tabular}
\par\smallskip
\begin{tabular}{@{}lrr@{}}
\toprule
Additional comparisons & A & D \\
\midrule
GR repair (\%, three-seed mean) & 49.34 & 53.78 \\
GR damage (\%, three-seed mean) & 18.99 & 23.45 \\
IR global pooled AUC (seed 0) & 0.7720 & 0.7683 \\
IR truth ranked above Base error (seed 0) & 218/309 & 219/309 \\
\bottomrule
\end{tabular}

\end{table}

\subsection{Contribution of the R4 increment}\label{sec:d-c}
Relative to C, adding the R4 increment improves mean joint success by 4.85 points under GR, 2.80 points at legacy matched IR FAR, and 2.16 points at legacy matched IR R4. These gains are positive in all three seeds at each operating point. At matched R4, D also accepts the Base wrong count less often in every seed. Pooled R4-versus-R3 AUC gains average 0.0074. Table~\ref{tab:dc} provides the paired intervals and per-seed results. Seed 0 has the smallest D--C gain under both legacy matching rules.

\subsection{Hard-positive and group-risk comparisons}\label{sec:controls}
Qwen3.5 E adds 326 Base-rejected truths and omits 107 R4 positives relative to D. E identifies hard positives with one verification query per annotated count. D's selection uses five open-count queries. The grouped training protocol uses the Base modes for all arms, so E adds the verification pass. Both arms use the same update budget and total extra weight.

Group DRO's mean weights for R1--R5 over the final 100 grouped updates are 2.54\%, 1.84\%, 20.74\%, 53.01\%, and 21.87\%. It gives the most weight to R4 while retaining substantial weight on false premises after a miscount.

The seed-0 comparison places D, E, and Group DRO within one joint success under default GR (Table~\ref{tab:controls}). At matched FAR, the differences are \(+0.65\) points for D--E (95\% interval \([-1.03,2.47]\)), \(+1.62\) for D--DRO (\([-0.33,3.82]\)), and \(+0.97\) for E--DRO (\([-1.29,3.37]\)). All three pairwise intervals include zero under both readouts.

\begin{table}[!tbp]
\caption{Controlled comparisons on Qwen3.5 COCO-ReM Dev, seed 0. Joint success counts use 309 fixed Base-error pairs. The matched IR threshold is selected in five image-grouped folds using the default A FAR as the budget. FAR entries are percentages.}
\label{tab:controls}
\centering\small
\begin{tabular}{@{}lrrrrr@{}}
\toprule
& \multicolumn{2}{c}{Default GR} & Default IR & \multicolumn{2}{c}{Matched-FAR IR}\\
\cmidrule(lr){2-3}\cmidrule(lr){5-6}
Training & Joint & FAR & Joint & Joint & FAR\\
\midrule
A & 117 & 20.92 & 115 & 115 & 20.97\\
D & 129 & 26.00 & 123 & 122 & 20.97\\
E & 128 & 24.56 & 121 & 120 & 20.97\\
Group DRO & 129 & 26.46 & 127 & 117 & 20.87\\
\bottomrule
\end{tabular}
\end{table}

\section{Paired effects and repair by seed}
Tables~\ref{tab:da} and~\ref{tab:dc} report the per-seed paired intervals. Table~\ref{tab:repair-detail} reports repair and damage by readout. Under GR, D corrects more Base errors than A in each seed, and it reverses more of Base's correct answers.

\begin{table}[!tbp]
\centering\small
\caption{Full method D versus equal-budget continuation A on the 309 fixed Base-error Dev images. The acceptance column gives achieved R4 for matched-R4 rows and overall FAR for other rows. Joint successes are image counts. Changes and paired 95\% intervals are in percentage points.}
\label{tab:da}
\begin{tabular}{@{}clrrrr@{}}
\toprule
Seed & Readout & A successes & D successes & Acceptance (\%) A $\to$ D & $\Delta J_{43}$ [95\% CI] \\
\midrule
0 & GR & 117 & 129 & 20.92 $\to$ 26.00 & +3.88 [0.92, 7.09] \\
0 & IR & 115 & 123 & 20.97 $\to$ 24.46 & +2.59 [0.00, 5.37] \\
0 & IR matched FAR & 115 & 122 & 20.97 $\to$ 20.97 & +2.27 [0.00, 4.65] \\
0 & IR matched R4 & 115 & 122 & 45.63 $\to$ 45.31 & +2.27 [0.00, 4.90] \\
1 & GR & 114 & 120 & 19.79 $\to$ 24.05 & +1.94 [-1.01, 5.25] \\
1 & IR & 103 & 121 & 19.38 $\to$ 22.62 & +5.83 [3.11, 8.86] \\
1 & IR matched FAR & 103 & 115 & 19.38 $\to$ 19.23 & +3.88 [1.60, 6.39] \\
1 & IR matched R4 & 103 & 113 & 42.72 $\to$ 43.37 & +3.24 [0.66, 5.79] \\
2 & GR & 116 & 124 & 20.41 $\to$ 24.97 & +2.59 [-0.31, 5.41] \\
2 & IR & 119 & 118 & 20.62 $\to$ 22.77 & -0.32 [-2.25, 1.56] \\
2 & IR matched FAR & 119 & 118 & 20.62 $\to$ 20.72 & -0.32 [-2.32, 1.58] \\
2 & IR matched R4 & 119 & 117 & 47.57 $\to$ 47.57 & -0.65 [-2.52, 1.01] \\
\bottomrule
\end{tabular}
\end{table}
\begin{table}[!tbp]
\centering\small
\caption{R4 increment D versus protection control C on the 309 fixed Base-error Dev images. The acceptance column gives achieved R4 for matched-R4 rows and overall FAR for other rows. Joint successes are image counts. Changes and paired 95\% intervals are in percentage points.}
\label{tab:dc}
\begin{tabular}{@{}clrrrr@{}}
\toprule
Seed & Readout & C successes & D successes & Acceptance (\%) C $\to$ D & $\Delta J_{43}$ [95\% CI] \\
\midrule
0 & GR & 117 & 129 & 18.87 $\to$ 26.00 & +3.88 [0.32, 7.52] \\
0 & IR & 115 & 123 & 20.15 $\to$ 24.46 & +2.59 [-0.32, 5.57] \\
0 & IR matched FAR & 115 & 121 & 20.15 $\to$ 20.21 & +1.94 [0.00, 4.19] \\
0 & IR matched R4 & 118 & 122 & 45.95 $\to$ 45.31 & +1.29 [-0.66, 3.37] \\
1 & GR & 106 & 120 & 18.46 $\to$ 24.05 & +4.53 [1.56, 7.75] \\
1 & IR & 103 & 121 & 18.26 $\to$ 22.62 & +5.83 [2.95, 8.81] \\
1 & IR matched FAR & 103 & 112 & 18.26 $\to$ 18.46 & +2.91 [0.35, 5.38] \\
1 & IR matched R4 & 107 & 114 & 44.98 $\to$ 46.28 & +2.27 [0.00, 4.65] \\
2 & GR & 105 & 124 & 18.77 $\to$ 24.97 & +6.15 [3.18, 9.21] \\
2 & IR & 105 & 118 & 19.54 $\to$ 22.77 & +4.21 [1.64, 6.82] \\
2 & IR matched FAR & 105 & 116 & 19.54 $\to$ 19.54 & +3.56 [1.58, 5.99] \\
2 & IR matched R4 & 109 & 118 & 45.63 $\to$ 46.28 & +2.91 [0.95, 5.18] \\
\bottomrule
\end{tabular}
\end{table}

\begin{table}[!tbp]
\centering\small
\caption{Repair and damage of frozen Base decisions on COCO-ReM Dev (\%). Repair uses 481 GR errors or 517 IR errors. Damage uses 1,969 correct GR conditions or 1,933 correct IR conditions.}
\label{tab:repair-detail}
\begin{tabular}{@{}lrrrrr@{}}
\toprule
Readout & Seed & A repair & D repair & A damage & D damage \\
\midrule
GR & 0 & 48.02 & 52.39 & 19.55 & 24.48 \\
GR & 1 & 50.31 & 55.30 & 18.44 & 22.50 \\
GR & 2 & 49.69 & 53.64 & 18.99 & 23.36 \\
IR & 0 & 49.52 & 52.22 & 19.35 & 22.81 \\
IR & 1 & 51.26 & 55.32 & 17.74 & 20.85 \\
IR & 2 & 51.84 & 53.00 & 18.83 & 20.85 \\
\bottomrule
\end{tabular}
\end{table}

\end{document}